\documentclass{report}
\usepackage[utf8]{inputenc}
\usepackage{nameref}
\usepackage[numbers]{natbib}
\usepackage{amsmath}
\usepackage{graphicx}
\usepackage{url}
\usepackage{listings}
\usepackage[section]{placeins}
\usepackage[T1]{fontenc}
\usepackage{geometry}
\usepackage{titlesec}
\titleformat{\chapter}[display]
  {\normalfont\huge\bfseries}{}{0pt}{\Huge}
\titlespacing*{\chapter}
  {0pt}{10pt}{40pt}

\usepackage{xcolor}

\definecolor{codegreen}{rgb}{0,0.6,0}
\definecolor{codegray}{rgb}{0.5,0.5,0.5}
\definecolor{codepurple}{rgb}{0.58,0,0.82}
\definecolor{codeteal}{rgb}{0.2, 0.6, 0.7}
\definecolor{backcolour}{rgb}{0.85,0.85,0.84}

\lstdefinestyle{mystyle}{
    backgroundcolor=\color{backcolour},   
    commentstyle=\color{codegreen},
    keywordstyle=\color{codeteal},
    numberstyle=\tiny\color{codegray},
    stringstyle=\color{codepurple},
    basicstyle=\small\color{black}\ttfamily,
    breakatwhitespace=false,         
    breaklines=true,                 
    captionpos=b,                    
    keepspaces=true,                 
    numbers=left,                    
    numbersep=7pt,                  
    showspaces=false,                
    showstringspaces=false,
    showtabs=false,                  
    tabsize=2
}
\title{Utilizing ROS 1 and the Turtlebot3 in a Multi-Robot System
\\
  \large A technical report of the Robotics and Automation Design Lab
}

\author{Corey Williams$^1$, Adam Schroeder$^1$}
\date{%
    $^1$Department of Mechanical, Industrial, and Manufacturing Engineering\\%
    The University of Toledo, Toledo, OH USA\\
    Corey.Williams3@rockets.utoledo.edu, Adam.Schroeder@utoledo.edu \\[2ex]%
    October 2020

}

\begin{document}

\maketitle
\tableofcontents
\pagebreak

\chapter*{Foreword} \label{Foreword}
\addcontentsline{toc}{chapter}{Foreword}
Our lab purchased our first turtlebots in September 2019 and had not used ROS before. The emanual documentation for the turtlebot3 from Robotis was excellent in getting us up and running. We encountered obstacles through that process, but were always able to overcome them. 

Next, we wanted to start making our turtlebots do things beyond what was covered in the Robotis documentation. For example, we wanted ROS to boot automatically when our turtlebot started up, and we wanted robots to automatically connect to our wireless router, and we wanted multiple robots to work using the same map. Accomplishing these tasks required us to pull together pieces of information from many different sources and much iterating.

We wrote this guide with the intention of supplementing the existing turtlebot documentation and with the hope of smoothing the way for beginners like us who want to work with multiple turtlebots. If you are a beginner, we would encourage you as you read to not be overwhelmed at the thought of trying to implement all of the ideas and tools that we present. Rather, we would recommend implementing them bit by bit, as needed.

Corey has performed the vast majority of the technical work, led the drafting of this document, and continues to use these turtlebots for his graduate research. I (Adam) encouraged Corey to create this guide, helped review, revise, and edit it along the way, and continue to direct research using turtlebots in the Robotics, Automation, and Design Lab (RADL) at the University of Toledo.

\chapter{Introduction} \label{Introduction}
\section{Purpose}
ROS (Robot Operating System) has become ubiquitous for testing new algorithms, alternative hardware configurations, and prototyping. By performing research with its modular framework, it can streamline sharing new work and integrations. However, it has many features and new terms that can take a considerable amount of time to learn for a new user. This paper will explore how to set up and configure ROS and ROS packages to work with a multi-robot system on a single master network.

\section{Scope}
As a prerequisite, it is important to know how to navigate a Linux based operating system and its command line, basic LAN configuration, and general programming skills. It is highly recommended to explore some of these topics if this is the first exposure. The primary goal here is to show how this knowledge can be used to configure existing systems to work with multiple robots along with introducing a few useful tools and not replacing the official Turtlebot3 eManual. Terminology will be explained where possible if obscure enough or specific to ROS. Users more familiar with Linux, the Turtlebot3, and ROS can skip to section 4 for information on configuring ROS.

\section{Organization}
The general overview can be viewed as:
\begin{itemize}
    \item [--] \nameref{TheTools}: a description of the hardware and software
    \item [--] \nameref{Linux}: configuring the OS and network for user-friendliness
    \item [--] \nameref{MRR}: configuring ROS for single master, multi-robot
    \item [--] \nameref{Usability}: optional topics to improve use
    \item [--] \nameref{Trouble}: how to use available tools to fix issues
\end{itemize}

\chapter{The Tools} \label{TheTools}
\section{ROS}
The ROS is a meta-operating system developed by Open Source Robotics Foundation \cite{osrf} that provides a modular framework for implementing robotics logic. Multiple ROS functionalities can be contained within a single folder called a package, and each functionality within that package can be broken down into separate scripts called nodes.  The power of ROS comes from how these nodes can communicate with each other; instead of directly passing data between each other, they can send a message to a topic that any number of nodes can subscribe to.  Consequently, each node does not need to know who is receiving the message, and a new node can replace an old one with better features without touching the rest of the system.

Challenges can arise if the same node is launched multiple times. Only one can be active, but there are tools provided by launch files, the file ROS utilizes to start programs, to launch more than one. For example, by adding a namespace (a prefix used to identify an instance of a node), a node can be renamed from $/camera\_ publisher$ to $/robot3/camera\_ publisher$. This allows for other nodes in a different namespace to be launched. Not everything will perform as intended without some work though. These issues are touched on in section 4. A workstation on the network needs to be running the naming and registration services contained in roscore \cite{master}. If this will not work with your application, ROS2 needs to be used.

\begin{figure}[h!]
\centering
\includegraphics[scale=0.5]{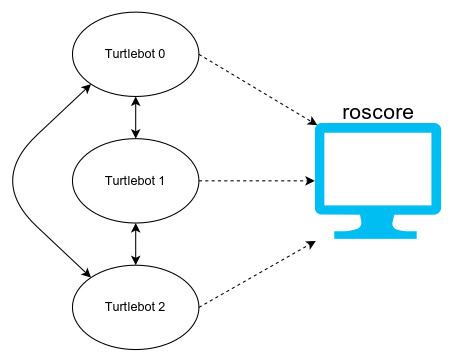}
\caption{ROS Network}
\label{fig:network}
\end{figure}

\subsection{Multi-Robot Systems}
A multi-robot system (MRS) is a system that contains more than one robot capable of communicating with one another. With some effort, it is possible to create emergent behaviors and active collaboration. There are two main categories of MRS: decentralized and centralized. As the names would suggest, a centralized framework is composed of a top-down command structure. With a decentralized system, each robot will take in information about its surroundings and possibly the states of its neighbor robots. Using a ruleset, it can decide on an action independent of any central authority. Variations of these can be done; a user taking momentary, manual control over a group could be an example. ROS 1 has some limitations on the decentralized side. The biggest issue is that node discovery is done through a single computer known as the master. Once two nodes are aware of each other and passing information through a topic, the connection is peer-to-peer. However, any changes in communication would require a connection to the master \cite{master}.
\section{Turtlebot3}
The Turtlebot3 is a modular robot platform sold by Robotis that includes enough hardware to begin working with autonomous 2D navigation. The work in this paper is based on the “Burger” model, but other models should follow the same logic. Much of the work to implement many ROS features, such as simulations, drivers, SLAM, and navigation, is handled in the Turtlebot3 packages \cite{robotispkg} \cite{robotissim}. These can serve as excellent examples on how to set up a launch file.
\subsection{ROS Packages}
Robotis maintains two repositories: one for controlling the robot \cite{robotispkg}, and one for running simulations \cite{robotissim}. The notable packages include bringup, navigation, slam, and teleop. Bringup will initiate the hardware and state publisher, navigation starts nodes needed to enable autonomous navigation, slam starts nodes needed for a SLAM operation, and teleop allows for manual control over the robot. Each one is considered its own package, but they are provided by a single root folder for convenience.
\subsection{LiDAR}
The primary sensor used for navigation is the 360° LiDAR. One of the most used SLAM algorithms, gmapping, utilizes the scan data published by the LiDAR. This should cover robot vision requirements so long as the environment can be navigated through a 2D occupancy grid, there are no obstacles out of the LiDAR’s planar line of sight, or slim obstacles that can’t reliably reflect the laser. 
\subsection{Single Board Computer}
Many robots can be controlled through a microcontroller, but ROS and the algorithms it will run require much more computational power. Current Turtlebot3 models provide a Raspberry pi 3 B+ as its single-board computer (SBC). It will act as the brain for each robot.
\subsection{OpenCR}
The OpenCR is an open-source controller board with many features. On the OpenCR board is an Arduino compatible microcontroller, gyroscope, accelerometer, several communication ports, and more \cite{opencr}. This board is where the power and motors are connected. The SBC also communicates and receives power through the micro USB port.

\chapter{Linux} \label{Linux}
\section{Why Use Linux}
Starting with the release of ROS Melodic, Windows is a supported platform, but most resources continue to write with Linux Distributions in mind, specifically with Ubuntu. The built-in package manager "apt" can handle the installation procedure once the ROS keyserver is added. In addition, both Ubuntu Server and Raspian are freely available to install on as many robots as desired and can run on the Raspberry Pi. Robotis provides a Raspian image with everything needed pre-installed; the download can be found in the SBC setup section of the e-Manual.
\section{Network Setup}
A quality network and network configuration are vital to get adequate performance with ROS. When all nodes are being run on a single machine and can communicate to the rest of the fleet through other means, only the localhost would be needed, and performance would be exceptional with minimal effort. Otherwise, every robot needs to be connected to a single network. The OS and ROS must be configured to connect properly.
\subsection{Network Connection on Linux}
In Ubuntu, the network connection can be controlled through the YAML based Netplan configuration file (Netplan). It is located at $/etc/wpa\_supplicant/wpa\_supplicant.conf$, and a simple setup suggested by Robotis \cite{emanual} is shown below:

\begin{lstlisting}[caption={Ubuntu Netplan}]
network:
  version: 2
  renderer: networkd
  ethernets:
    eth0:
    optional: true
    dhcp4: yes
    dhcp6: yes
  wifis:
    wlan0:
    dhcp4: yes
    dhcp6: yes
    access-points:
      "network-ssid":
        password: "network-password"
\end{lstlisting}

Keep in mind that YAML is like Python in that the whitespace indicates how the script is parsed. While the e-manual specified this for a machine running ROS 2, where static IP’s are not necessary, it is possible to control the IP addresses of the fleet from the router interface by enabling DHCP on the robot or workstation level.

When setting up a Raspberry Pi in a headless manner, it is possible to add the $wpa\_supplicant.conf$ file to the boot folder on the SD card before the initial boot. These settings should be automatically added to the correct folder, and no direct intervention is necessary. The Raspberry Pi foundation provides a minimal setup script \cite{raspwireless}. Note that as of this writing, the Robotis provided distribution listed in the Turtlebot3 e-manual does not function properly with this method:

\begin{lstlisting}[caption={Headless Raspberry Pi Setup}]
ctrl_interface=DIR=/var/run/wpa_supplicant GROUP=netdev
update_config=1
country=<Insert 2 letter ISO 3166-1 country code here>
network={
 ssid="<Name of your wireless LAN>"
 psk="<Password for your wireless LAN>"
}

\end{lstlisting}

\subsection{ROS Network Variables}
ROS utilizes TCP/IP sockets for sending and receiving data between nodes \cite{tcp}. This means that any home or industrial class router can be used. If a single robot is operating with no external workstation as a master, ROS can operate through the localhost, and no networking hardware is required. There are two environment variables necessary to get communication working: $ROS\_HOSTNAME$ and $ROS\_MASTER\_URI$ \cite{netsetup}. If the IP of the master is 192.168.0.2 and the robot is 192.168.0.3, this can be done in the $.bashrc$ file with 2 additional lines.

\begin{lstlisting}[language=bash, caption={Network Addresses}]
$ export ROS_HOSTNAME=192.168.0.3
$ export ROS_MASTER_URI=http://192.168.0.2:11311
\end{lstlisting}

The 11311 at the end of the master IP indicates the port number. If DHCP is enabled, these numbers could become outdated. Assigning IP’s through the router can prevent this and provide a central place to make future modifications. It is possible to use mDNS to not require static IP’s and provide an easier to remember naming scheme. Ubuntu, and other Linux distributions, can utilize Avahi to handle this automatically. Instead of knowing the IP, it is possible to resolve $hostname.local$ to the IP belonging to a computer with that hostname. In a Debian based OS it can be installed with:

\begin{lstlisting}[language=bash, caption={Installing Avahi}]
$ sudo apt install avahi-daemon
\end{lstlisting}

OSX comes with Bonjour and is compatible with Avahi \cite{avahi}. Windows needs either iTunes or the Bonjour print service installed to do the same. The print service is a small program that can perform the setup needed \cite{applebonjour}. While this can be convenient, if the mDNS service ever stops functioning, that machine can no longer parse or advertise $.local$ addresses. The new environment variables will look like the following:

\begin{lstlisting}[language=bash, caption={Avahi Addresses}]
$ export ROS_HOSTNAME=hostname.local
$ export ROS_MASTER_URI=http://masterhost.local:11311
\end{lstlisting}

\section{Utilizing SSH}
It would be unreasonable to connect every robot to a keyboard and monitor when changes need to be made. Instead, run terminal commands remotely using SSH. Most commonly, OpenSSH is used, and the client and server can be installed through the command:

\begin{lstlisting}[language=bash, caption={Installing SSH}]
$ sudo apt install openssh-client openssh-server
\end{lstlisting}

To use it, open a new terminal to take place of a terminal on another machine. Then, assuming you are connecting from the master to the robot, type:
\begin{lstlisting}[language=bash, caption={Utilizing SSH}]
$ ssh username@192.168.0.3
# If mDNS is used:
$ ssh username@hostname.local
\end{lstlisting}

Upon first connect, include the option $-oHostkeyAlgorithms=`ssh-rsa\textrm'$; this is important if remote node calls need to be made by using launch files. It will ask whether you wish to add the device to known devices or not. Enter yes, and the password to the user will be prompted. It is not a good idea to have every device use the same password for security reasons but remembering every password and entering one every time is not convenient. To get around this, it is possible to generate SSH keys that can be copied over to the client. Raspberry Pi foundation has instructions on their website \cite{raspssh}, but the basic steps for a first time use are listed here to summarize:

\begin{enumerate}
    \item $ssh-keygen$ if no keys exist
    \item $ssh-copy-id$ $username@ip\-address$
    \item Sign in when prompted
\end{enumerate}

\section{Synchronizing Time}
Some communications will have problems if a time discrepancy exists between different machines on the network. ROS includes instructions on how to prevent this by using Chrony \cite{netsetup}. Chrony is an NTP based time management software; importantly, it shares compatibility with Linux and macOS \cite{chrony}. With this, it is possible to configure one of the machines on a network as a time server where the other machines will obtain the current time. OSRF provides a recommendation for adding a server to the $/etc/chrony/chrony.conf$ configuration file \cite{netsetup}. Using the IP address naming convention from earlier, on the client add the line:

\begin{lstlisting}[caption={Client Chrony Config Line}]
server 192.168.0.2 minpoll 0 maxpoll 5 maxdelay .05
\end{lstlisting}

By not removing the existing internet-based servers, the Raspberry Pi can properly obtain a time if it ever needs to use the internet again. On the server add the following line:
\begin{lstlisting}[caption={Server Chrony Config Line}]
allow 192.168.0.0/24
\end{lstlisting}

With both .conf files saved, check to ensure that the system is operating with the two commands:

\begin{lstlisting}[language=bash, caption={Force Time Synchronization}]
$ sudo chronyc -a makestep
$ chronyc tracking
\end{lstlisting}

A readout of the performance will be printed in the terminal. The “Leap Status” row will specify if everything is working normally. To check if the client has the correct server use this command:

\begin{lstlisting}[language=bash, caption={Check Registered Servers}]
$ chronyc sources
\end{lstlisting}

{\let\clearpage\relax \chapter{Multi-Robot ROS}\label{MRR}}
\section{Simulation}
Being able to simulate of a robot with the same code that would be placed onto the real model can make iteration and debugging easier. Hardware issues such as battery charge or networking setup can be eliminated. Robotis provides packages for use in the Gazebo simulator \cite{robotissim}. Gazebo can emulate a robot with features such as cameras, encoders, manipulators, and TCP/IP communication; however, it will be replaced by Ignition Gazebo when Gazebo Classic reaches end-of-life in 2025 \cite{ignition}. Many other ROS compatible robot manufacturers also include packages for Gazebo.
To begin, we will deconstruct the $multi\_turtlebot3.launch$ XML file included in the $turtlebot3\_gazebo$ package \cite{robotissim}. This will be a great starting point when making a new one from scratch with either a Turtlebot3 or something else. It can be broken into 3 sections. The first can be adequately reduced to:

\begin{lstlisting}[language=html, caption={First Turtlebot Setup}]
<arg name="model" default="$(env TURTLEBOT3_MODEL)" doc="model type [burger, waffle, waffle_pi]"/>
<arg name="first_tb3"		default="tb3_0"/>
<arg name="second_tb3"	default="tb3_1"/>
<arg name="third_tb3"		default="tb3_2"/>

<arg name="first_tb3_x_pos"	default="-7.0"/>
<arg name="first_tb3_y_pos"	default="-1.0"/>
<arg name="first_tb3_z_pos"	default=" 0.0"/>
<arg name="first_tb3_yaw"   default=" 1.57"/>
\end{lstlisting}

Each element here indicates an argument that can be passed through the command line, where the name attribute is the name of the argument.  The first argument in the list indicates the model of Turtlebot3 being used and pulls in the environment variable $TURTLEBOT3\_MODEL$ as the default value. The next three specify the namespace of each robot; this will group the scripts and topics associated with a single robot together. The rest indicate a pose (x, y, z, and z-axis rotation) of each robot to be placed within Gazebo. The next section consists of launching the desired gazebo world.

\begin{lstlisting}[language=html, caption={Launching A Gazebo World}]
<include file="$(find gazebo_ros)/launch/empty_world.launch">
    <arg name="world_name"    value="$(find turtlebot3_gazebo)/worlds/turtlebot3_house.world"/>
    <arg name="paused"        value="false"/>
    <arg name="use_sim_time"  value="true"/>
    <arg name="gui"           value="true"/>
    <arg name="headless"      value="false"/>
    <arg name="debug"         value="false"/>
</include>  
\end{lstlisting}
The include element tells the launch file to run another launch file. Instead of providing the path to the package directly, which may change depending on where the user has it installed, the $\$(find\; pkg)$ shortcut returns the path to the package root folder. From there it is possible to provide the path to the desired launch file. The $arg$ sub-elements will be passed to the included launch file. The important one to look at here is the $world\_name$ argument. When a gazebo $.world$ file is provided, it will automatically be loaded when the launch file is run. For the last section, the robots are initialized and spawned into the Gazebo world.

\begin{lstlisting}[language=html, caption={Launching Core Turtlebot Nodes}]
<group ns = "$(arg first_tb3)">
    <param name="robot_description" command="$(find xacro)/xacro --inorder $(find turtlebot3_description)/urdf/turtlebot3_$(arg model).urdf.xacro" />

    <node pkg="robot_state_publisher" type="robot_state_publisher" name="robot_state_publisher" output="screen">
        <param name="publish_frequency" type="double" value="50.0" />
        <param name="tf_prefix" value="$(arg first_tb3)" />
    </node>
    
    <node name="spawn_urdf" pkg="gazebo_ros" type="spawn_model" args="-urdf -model $(arg first_tb3) -x $(arg first_tb3_x_pos) -y $(arg first_tb3_y_pos) -z $(arg first_tb3_z_pos) -Y $(arg first_tb3_yaw) -param robot_description" />
</group>
\end{lstlisting}

As the group element name suggests, each of the elements above are grouped under the $first\_tb3$ namespace. Instead of specifying the $ns$ tag in every relevant element, grouping them together can handle this with only 2 additional lines; this can also be helpful when using if statements, which will be covered in section 6.3: Recursive Launch Files. The $\$(arg\: first\_tb3)$ line recalls the value passed by the $first\_tb3$ argument specified at the top of the launch file.

Many of the required nodes for a Turltebot3, such as $turtlebot3\_core$, are handled by Gazebo when spawning a robot instance. The above XML can be pasted for as many robots as needed. The only modifications needed are to the arguments for the namespace, tf prefix, and starting position.
\section{Bringup}
There are two main launch files important to the user in the \textit{turtlebot3\_bringup} package: \textit{turtlebot3\_robot.launch} and \textit{turtlebot3\_remote.launch}. \textit{Turtlebot3\_robot} will initiate the hardware and diagnostics. \textit{Turtlebot3\_remote} handles starting the \textit{robot\_state\_publisher} node. Starting these nodes with multiple robots is very straight forward. Add \textit{ROS\_NAMESPACE=[robot name]} in the beginning of the command and pass the desired robot namespace to the \textit{multi\_robot\_name} argument when launching the “robot” and “remote” bringup scripts. More information can be found in section 15.5 of the Turtlebot3 e-Manual \cite{emanual}.
\section{SLAM}
Although it is possible, multi-robot SLAM is not conventionally done. Instead, a single robot is launched to map the desired area and the results are saved using the \textit{map\_server} package. \textit{Map\_server} has functions to both save and load maps as a \textit{.pgm} file and an accompanying \textit{.yaml} data file \cite{amcl}. The Turtlebot3 e-Manual suggests the \textit{multi\_map\_merge} package \cite{emanual}, but it has not been updated since Kinetic \cite{horner}. More work has been done in the realm of collaborative mapping; Google has its own solution using gRPC and Cartographer, but its documentation is not completed \cite{google}.

To begin, follow the instructions in the e-Manual in section 9. By the end there should be a map saved to your drive. Depending on the results, most features from the real or virtual environment should be recognizable in the top-down view. It is possible to take advantage of this fact to manually remove mapping errors. The free and open-source photo editor GIMP is capable of editing \textit{.pgm} files. Using the pencil tool, manually change pixels to white (\#fefefe), black (\#000000), or gray (\#cdcdcd) depending on whether the space is free, occupied, or unknown respectively. If desired, paths that are open can be blocked by adding black pixels.

\begin{figure}[h!]
    \centering
    \begin{minipage}{0.45\textwidth}
        \centering
        \includegraphics[scale=0.95]{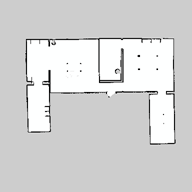} 
        \caption{Raw SLAM Map}
    \end{minipage}\hfill
    \begin{minipage}{0.45\textwidth}
        \centering
        \includegraphics[scale=0.95]{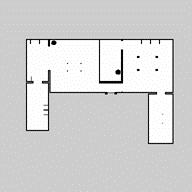} 
        \caption{Modified SLAM Map}
    \end{minipage}
\end{figure}

\section{AMCL}
The map provided by the map server acts as a unified coordinate system that all the robots can localize themselves too. This localization is done with a package called AMCL (adaptive Monte Carlo localization); it uses a particle filter to estimate the current pose of the robot. Once an estimated is obtained, it will handle the transform from the \textit{/map} to \textit{/robot\_name/odom} reference frames.

Launch files associated with amcl can involve a lot of parameters; copying the launch file AMCL.launch from the \textit{turtlebot3\_navigation} package into your own package can speed up the process.
\pagebreak
\begin{lstlisting}[language=html, caption={Launching AMCL Under A Namespace}]
<!-- Argument to be added to the top of the launch file -->
<arg name="multi_robot_name" default="" doc="User assigned robot namespace."/>

<!-- Existing lines with robot name references in red -->
<group ns="$(arg multi_robot_name)">
   <node pkg="amcl" 	type="amcl"	name="amcl">
      <remap from="scan"		to="$(arg multi_robot_name)/$(arg scan_topic)"/>
      <remap from="initialpose"		to="$(arg multi_robot_name)/initialpose"/>
      <remap from="amcl_pose"		to="$(arg multi_robot_name)/amcl_pose"/>
      <remap from="particlecloud"	to="$(arg multi_robot_name)/particlecloud"/>

      <param name="odom_frame_id"	value="$(arg multi_robot_name)/odom"/>
      <param name="base_frame_id"	value="$(arg multi_robot_name)/base_footprint"/>
   </node>
</group>
\end{lstlisting}

By providing the \textit{multi\_robot\_name} argument, all relevant topics are changed to work with that specific robot while still providing a transform to the common global frame of reference \textit{/map}.

\section{Map Server}
The map server is a node that can take \textit{.yaml} and \textit{.pgm} files to provide for the rest of the network \cite{mapserver}. It is different from many other nodes that will be launched; in our use case, it does not care how many robots are active and will not require a name space. If only one map file is needed, only one node is needed. Each robot will obtain the map from the \textit{static\_map} service call by default. Alternatively, the map can be pulled from the /map and \textit{/map\_metadata} topics by setting the \textit{use\_map\_topic} \textit{amcl} argument to true \cite{amcl}. The map is loaded by passing the \textit{.yaml} file through the \textit{args} attribute like below if the map and metadata are located located in a maps folder in the package.

\begin{lstlisting}[language=html, caption={Launching A Map Server}]
<node	pkg="map_server" name="map_server" type="map_server" args="$(find pkg_name)/maps/$(arg map_name).yaml"/>
\end{lstlisting}

\section{Move Base}
The \textit{move\_base} package is the path planning algorithm that comes with ROS. There are many other supporting packages, but \textit{move\_base} provides the \textit{MoveBaseAction} service to send navigation targets to \cite{movebase}.  To set up for a multi-robot environment, create a launch file similar to the one below. Instances specific to taking namespaces into account are highlighted in red for emphasis.
\pagebreak
\begin{lstlisting}[language=html, caption={Launching Move Base Under A Namespace}]
<!-- Argument to be added to the top of the launch file -->
<arg name="multi_robot_name"/>

<group ns="$(arg multi_robot_name)">
    <node pkg="move_base" type="move_base" respawn="false" name="move_base" output="screen">
	
      <!-- Set tf_prefix for frames explicitly, overwriting defaults -->
      <param name="global_costmap/scan/sensor_frame" value="$(arg multi_robot_name)/base_scan"/>
      <param name="global_costmap/obstacle_layer/scan/sensor_frame"
	     value="$(arg multi_robot_name)/base_scan"/>
      <param name="global_costmap/global_frame" value="map"/>
      <param name="global_costmap/robot_base_frame"
	     value="$(arg multi_robot_name)/base_footprint"/>
      <param name="local_costmap/scan/sensor_frame" value="$(arg multi_robot_name)/base_scan"/>
      <param name="local_costmap/obstacle_layer/scan/sensor_frame"
	     value="$(arg multi_robot_name)/base_scan"/>
      <param name="local_costmap/global_frame" value="$(arg multi_robot_name)/odom"/>
      <param name="local_costmap/robot_base_frame"
	     value="$(arg multi_robot_name)/base_footprint"/>

      <!-- Centralized map server -->
      <remap from="map" to="/map"/>
    </node>
  </group>

\end{lstlisting}
\pagebreak
\section{RVIZ}
Like Gazebo, RVIZ can be loaded using a launch file. The only requirement to begin is to have a \textit{.rviz} file available. If there is a configuration that is required but not provided by another packages, make any changes in a RVIZ window and save with:

\begin{equation*}
    \textrm{File} \rightarrow \textrm{Save Config As}
\end{equation*}

Save the file somewhere convenient; this will most likely be going to be somewhere in your package. In this instance, let us assume that it was saved in a folder called \textit{rviz} located in your package’s root folder. Then, in a launch file, include this line:

\begin{lstlisting}[language=html, caption={Loading RVIZ File From A Launch File}]
<node pkg="rviz" type="rviz" name="rviz" args="-d $(find your_package)/rviz/your_file.rviz"/>
\end{lstlisting}

\begin{figure}[ht]
\centering
\includegraphics[scale=0.95]{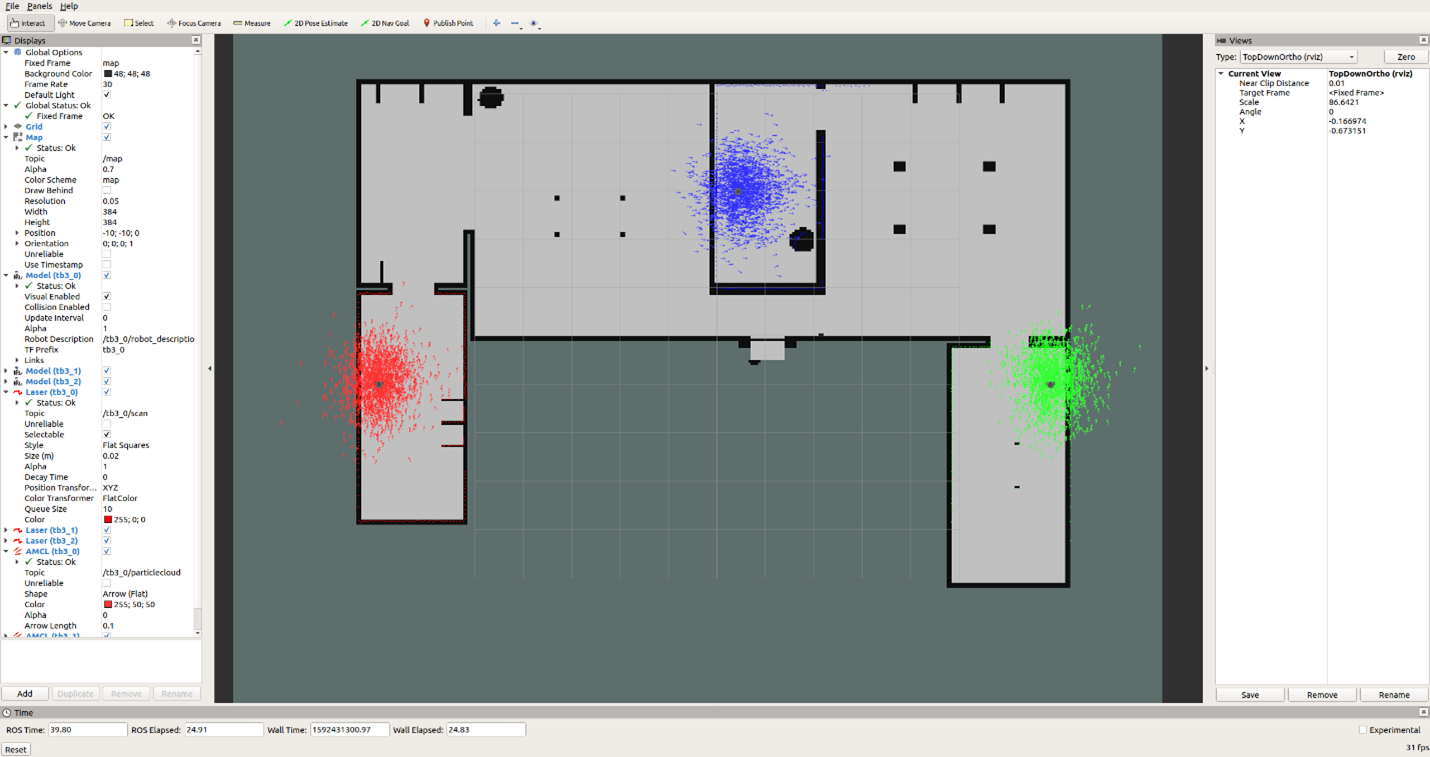}
\caption{RVIZ With 3 Turtlebots}
\label{fig:rviz}
\end{figure}

\chapter{Usability Improvements} \label{Usability}
\section{Combining Launch Files}
Launching each of these components for a single robot may not take much effort, but it’s not a desirable solution when more are added. Several steps can be combined into one with the include tag demonstrated in section 4.1 Simulation. It may be helpful to start with a process flow map. Make each step its own launch file that can be confirmed to function as intended; then, create an extra launch file that can unify all the other steps with a single command.
\begin{figure}[ht]
\centering
\includegraphics[scale=0.5]{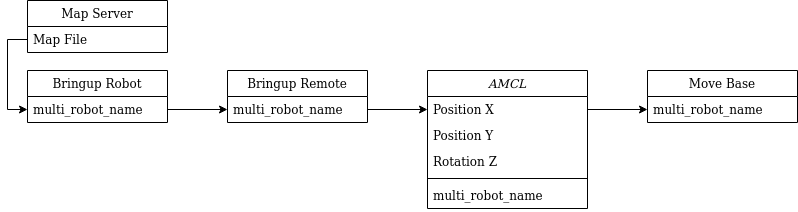}
\caption{RVIZ With 3 Turtlebots}
\label{fig:launch}
\end{figure}

Now that each step has a launch file it is important to decide how the process is going to function. For example, only one map server is needed; it would not make sense to include it when launching a single robot. When the next one is launched, another map server will be started and the old one will be shut down. \textit{Bringup\_robot} is typically ran using the Raspberry Pi’s hardware and the rest is done on a more powerful workstation. By taking this into account, a few judgement calls can be made. \textit{Roscore} and \textit{map\_server} can be launched once on the workstation without a problem, and the rest can be launched by the robots on boot and ran where needed by defining machines in the launch file. This process is outlined with the launch file  fragment here: 
\pagebreak
\begin{lstlisting}[language=html, caption={Combination Launch File}]
<launch> 
  <!-- REQUIRED ARGUMENTS --> 
  <arg name="model"               default="$(env TURTLEBOT3_MODEL)"/> 
  <arg name="multi_robot_name"    default=""/> 
  <arg name="set_lidar_frame_id"  default="base_scan"/> 
  <arg name="user"  default="$(env WORKSTATION_USER)" doc="The username of the machine to offload to."/> 
  <arg name="address"  default="$(env WORKSTATION_ADDRESS)"/> 
  <arg name="target_version"  default="melodic"doc="The version of ros used on the machine to offload to."/> 

  <!-- WORKSTATION MACHINE DEFINITION --> 
  <machine name="workstation"     address="$(arg address)" env-loader="/opt/ros/$(arg target_version)/env.sh" user="$(arg user)"/> 

  <!-- TO BE CALLED ON ROBOT --> 
  <include file="$(find turtlebot3_bringup)/launch/turtlebot3_robot.launch"> 
    <arg name="multi_robot_name"    value="$(arg multi_robot_name)"/> 
    <arg name="set_lidar_frame_id"  value="$(arg set_lidar_frame_id)"/> 
  </include> 

  <!-- TO BE CALLED ON WORKSTATION --> 
  <arg name="urdf_file" default="$(find xacro)/xacro --inorder '$(find turtlebot3_description)/urdf/turtlebot3_$(arg model).urdf.xacro'"/> 
  <param name="robot_description" command="$(arg urdf_file)"/> 
  <node pkg="robot_state_publisher" type="robot_state_publisher" name="robot_state_publisher" machine="workstation"> 
    <param name="publish_frequency" type="double" value="50.0" /> 
    <param name="tf_prefix" value="$(arg multi_robot_name)"/> 
  </node> 
</launch> 
\end{lstlisting}

Notice that despite having an existing launch file for \textit{robot\_state\_publisher} that could be included, the contents were copied into this launch file. The machine attribute is not part of the include element; a node call must be used. Also, note that the ROS version does not necessarily have to be the same on both machines if the same code can be launched without changes on each. Python will likely have the most luck with this. In this case, the workstation was running Melodic while the Turtlebot3 was running Kinetic; however, it is necessary to have the package installed on whatever machine is making the remote call. Be sure to either include the password in the machine description (not recommended) or have SSH keys setup from the Turtlebot to the workstation. Launching AMCL and \textit{move\_base} can remain an exercise. 

\section{Daemons}

Now that everything is set up to launch with a few simple commands, it's possible to have the process automatically started on boot through a Daemon. Daemon is the name of a process that runs in the background; when it comes to Ubuntu and Raspian, daemons are handled by Systemd. Two files are needed for this to work: a shell script with the commands to launch the ROS nodes and a service file to register the process into Systemd. More advanced service files can be found in the write up by Rover Robotics; on their website, a page outlining starting \textit{roscore} and several nodes can be found \cite{robotics2019}. The following bash file can be used to launch the Turtlebot bringup:
\pagebreak

\begin{lstlisting}[language=bash, caption={Launch Script}]
#!/bin/bash 
source /opt/ros/[distro]/setup.bash 
source [directory-to-workspace]/devel/setup.bash 
export ROS_MASTER_URI = https://[roscore-ip-address]:11311 
export ROS_HOSTNAME = host-ip 
export ROS_NAMESPACE=[robot-name] 
roslaunch turtlebot3_bringup turtlebot3_robot.launch multi_robot_name:=[robot-name]
\end{lstlisting}

Save this shell script somewhere relevant. Rover Robotics uses \textit{/usr/sbin/roslaunch}.  Then make the script executable. This can be done by right clicking the file, clicking properties, and checking “allow executing file as program” in Ubuntu, setting execute to “anyone” in Raspian, or, if the terminal is desired, by using this command:

\begin{lstlisting}[language=bash, caption={Enabling Execution}]
$ chmod +x [filename].sh 
\end{lstlisting}

Now a \textit{.service} file is needed in \textit{/etc/systemd/system}. When making one, keep in mind any other services that it may need. Here we are waiting for the network to be online, the time to sync. Additionally, Avahi is used to allow the shell script to use \textit{.local} addresses instead of an IP address.

\begin{lstlisting}[caption={Service File}]
[Unit] 
After=network-online.service time-sync.target avahi-daemon.service

[Service] 
Type=simple 
User=[USERNAME] 
ExecStart=/[directory-to-launch-script]/[launch-script].sh 
Restart=on-failure 
RestartSec=3s 

[Install] 
WantedBy=multi-user.target 
\end{lstlisting}

On the chosen master, \textit{roscore} needs to be running before the robot is turned on. Otherwise, the node will fail to launch. 
\pagebreak
\section{Recursive Launch Files}
On some occasions it might be useful to launch an arbitrary number of nodes through a single launch file. This can be done through recursion and the \textit{eval} substitution argument. This can be illustrated here: 

\begin{lstlisting}[language=html, caption={0 Indexed Recursive Launch File}]
<launch> 
  <arg name="num"    default="1"/> 
  <include file="$(find pkg_name)/launch/node_instance.launch"> 
    <arg name="multi_robot_name"    value="$(eval str(arg('num') - 1))"/> 
  </include> 
  <include file="$(find pkg_name)/launch/arbitrary_launcher.launch" if="$(eval arg('num') - 1 > 0)"> 
    <arg name="num"    value="$(eval arg('num') - 1)"/> 
  </include> 
</launch> 
\end{lstlisting}

To begin, we will look at the \textit{num} argument; this is the number of nodes that needs to be launched. The next block is the launch file created to launch a specific node or a set of nodes, and the argument \textit{multi\_robot\_name} is passed to it. Its value is one less than the argument initially passed to make this 0-indexed. The next block will check if the value is greater than 0 if one is subtracted from it through the if attribute. It will then call itself again but pass a value to the \textit{num} argument as one less than it was when it started. The result is the launch file being called as many times as is passed to \textit{num}. 

{\let\clearpage\relax \chapter{Troubleshooting}\label{Trouble}}
Troubleshooting and debugging within ROS can be more complicated than other programming methods. Some algorithms could be completely functional but do not perform anything and do not report any errors. This has to do with how each node is configured. Luckily, ROS has visual and terminal based diagnostic tools. The list presented here is not exhaustive but should cover most issues. For more information, the ROS website will contain documentation on each tool \cite{rqtcommon}.
\section{Visual Diagnostic Tools}
The most common visual debugger is the node graph. In any available terminal with ROS sourced, this graph can be generated through the command:

\begin{lstlisting}[language=bash, caption={Viewing The Node Graph}]
$ rqt_graph
\end{lstlisting}

\begin{figure}[h]
\centering
\includegraphics[scale=0.22]{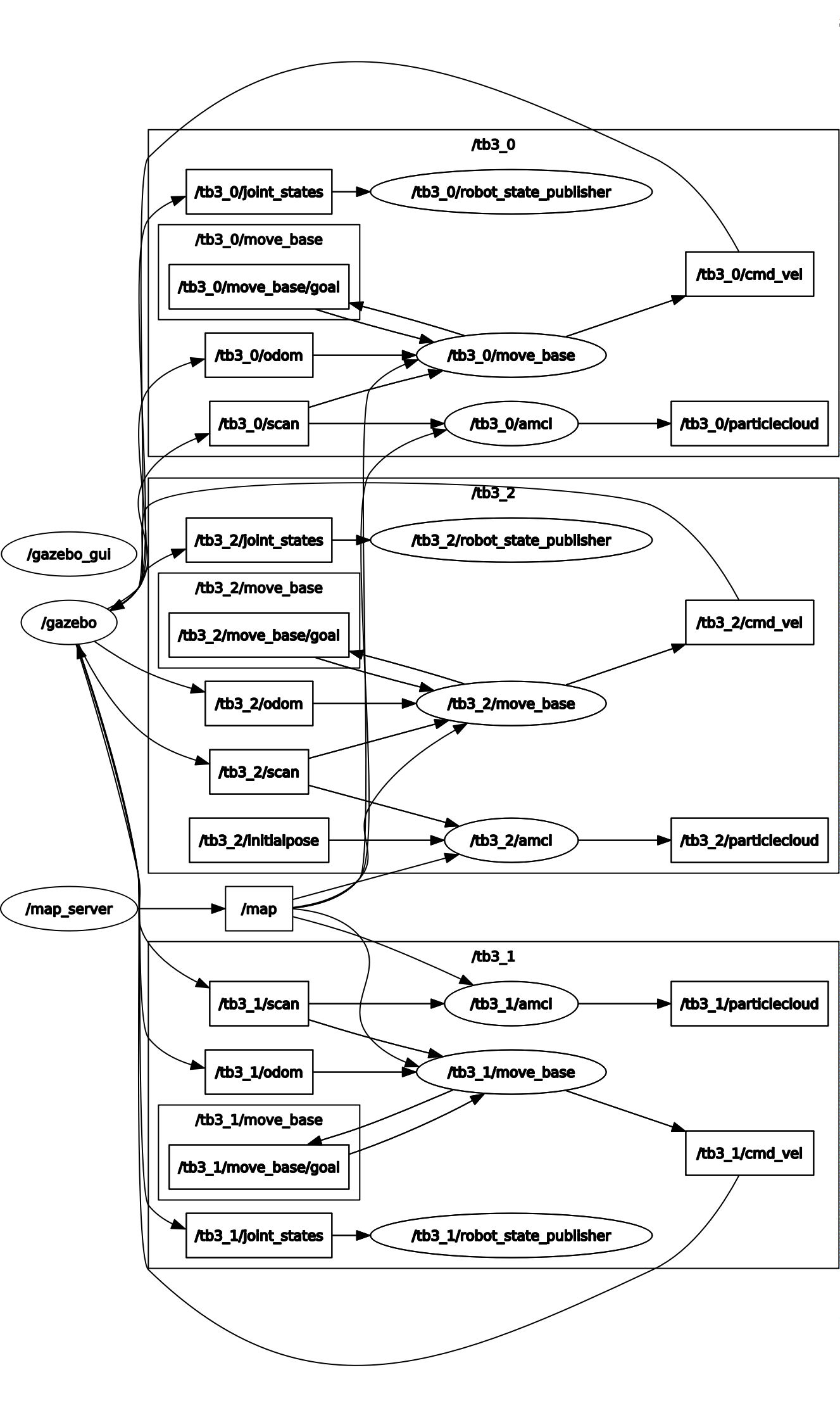}
\caption{Node Graph}
\label{fig:nodegraph}
\end{figure}

A window will then appear with each node and their connections through topics will be plotted. It can be helpful to look at the problem node first and follow the information stream until any faults can be found. There is an analogous version for the transform tree:

\begin{lstlisting}[language=bash, caption={Viewing The TF Tree}]
$ rosrun rqt_tf_tree rqt_tf_tree
\end{lstlisting}

\begin{figure}
\centering
\includegraphics[scale=0.075]{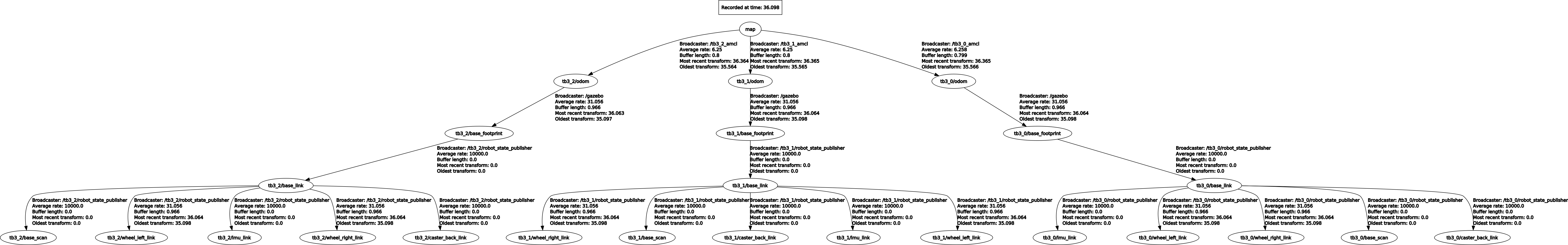}
\caption{TF Graph: Notice The Single Map Frame At The Top}
\label{fig:tfgraph}
\end{figure}

Diagnosing the problems inside a topic can be done through the plotter. Its UI is brought up with:
\begin{lstlisting}[language=bash, caption={Graphing Topic Information}]
$ rqt_plot
\end{lstlisting}

Topics can be added through the UI, and the graph window will begin to plot message data over time. This could bring about a discovery in data related issues. 

\section{Terminal Diagnostic Tools}
Occasionally, data cannot be plotted, or a simple glance at what is happening is sufficient. This can be done by using:

\begin{lstlisting}[language=bash, caption={Printing Topic Information To Terminal}]
$ rostopic echo /topic_name
\end{lstlisting}

That terminal will then print out the topic information as fast as it receives it but does not give any way to analyze it programmatically. \textit{Rosbag} is a good alternative to record these readouts to be played back or analyzed later. 
Some packages include debug information that is not visible without changing settings through the window called by:

\begin{lstlisting}[language=bash, caption={Viewing Debug Information}]
$ rosrun rqt_logger_level rqt_logger_level 
\end{lstlisting}

Here it is possible to select a node and different loggers located in its code; by changing the reporting level to “debug”, more information about how that node is functioning can be found. In rare circumstances, a malfunctioning node will instruct the user to change specific loggers to debug. 

\bibliographystyle{unsrt}
\bibliography{references}
\end{document}